%% file: sn-article.tex
\documentclass[sn-mathphys-num]{sn-jnl}

\usepackage{graphicx}%
\usepackage{multirow}%
\usepackage{amsmath,amssymb,amsfonts}%
\usepackage{array}
\usepackage{amsthm}%
\usepackage{mathrsfs}%
\usepackage[title]{appendix}%
\usepackage{xcolor}%
\usepackage{textcomp}%
\usepackage{manyfoot}%
\usepackage{booktabs}%
\usepackage{algorithm}%
\usepackage{algorithmicx}%
\usepackage{algpseudocode}%
\usepackage{listings}%
\usepackage{tabularx}

\theoremstyle{thmstyletwo}%

\theoremstyle{thmstylethree}%

\begin{document}

\title[Touching the tumor boundary: A pilot study on ultrasound based virtual fixtures for breast-conserving surgery]{Touching the tumor boundary: A pilot study on ultrasound based virtual fixtures for breast-conserving surgery 

}


\author*[1]{\fnm{Laura} \sur{Connolly}}\email{laura.connolly@queensu.ca}

\author[1]{\fnm{Tamas} \sur{Ungi}}

\author[2]{\fnm{Adnan} \sur{Munawar}}

\author[2]{\fnm{Anton} \sur{Deguet}}

\author[1]{\fnm{Chris} \sur{Yeung}}

\author[2]{\fnm{Russell H.} \sur{Taylor}}

\author[1]{\fnm{Parvin} \sur{Mousavi}}

\author[1]{\fnm{Gabor} \sur{Fichtinger}}

\author[1]{\fnm{Keyvan} \sur{Hashtrudi-Zaad}}

\affil*[1]{\orgname{Queen's University}, \orgaddress{ \city{Kingston}, \state{Ontario}, \country{Canada}}}

\affil[2]{\orgname{Johns Hopkins University},  \city{Baltimore}, \state{Maryland}, \country{United States}}

\abstract{\noindent\textit{This is an un-refereed author version of an article. 
The final refereed version is available here: 
\url{https://link.springer.com/article/10.1007/s11548-025-03342-z}}

\vspace{1em} 

\textbf{Purpose:} Delineating tumor boundaries during breast-conserving surgery is challenging as tumors are often highly mobile, non-palpable, and have irregularly shaped borders. To address these challenges, we introduce a cooperative robotic guidance system that applies haptic feedback for tumor localization. In this pilot study, we aim to assess if and how this system can be successfully integrated into breast cancer care.
\textbf{Methods:} A small haptic robot is retrofitted with an electrocautery blade to operate as a cooperatively controlled surgical tool. Ultrasound and electromagnetic navigation are used to identify the tumor boundaries and position. A forbidden region virtual fixture is imposed when the surgical tool collides with the tumor boundary. We conducted a study where users were asked to resect tumors from breast simulants both with and without the haptic guidance. We then assess the results of these simulated resections both qualitatively and quantitatively.  
\textbf{Results:} Virtual fixture guidance is shown to improve resection margins. On average, users find the task to be less mentally demanding, frustrating, and effort intensive when haptic feedback is available. We also discovered some unanticipated impacts on surgical workflow that will guide design adjustments and training protocol moving forward. 
\textbf{Conclusion:} Our results suggest that virtual fixtures can help localize tumor boundaries in simulated breast-conserving surgery.  Future work will include an extensive user study to further validate these results and fine-tune our guidance system.}

\keywords{Breast-conserving surgery, cooperative robotics, ultrasound, haptic feedback, virtual fixture.}

\maketitle

\section{Introduction}\label{sec1}

Approximately 7 in 10 patients diagnosed with early-stage breast cancer will undergo breast-conserving surgery (BCS) combined with radiation therapy to treat their disease \cite{christiansen2022breast}. During BCS, the surgeon will resect the tumor and a small portion of the surrounding healthy tissue in every direction. To determine if the resection was adequate, a pathologist will analyze the excised specimen and look at the distance between the edge of the specimen and the tumor. If this margin is considered positive (\textless{} 1 or 2 mm), it is implied that the tumor was transected during surgery and residual tumor tissue may be left behind \cite{bundred2022margin}. In these cases, the patient must undergo immediate reoperation, which can lead to additional complications, impaired cosmesis, increased healthcare costs, and psychological distress \cite{tamburelli2020reoperation}. Currently, between 10-40\% of patients who undergo BCS experience positive margins \cite{de2018tumor}\cite{racz2020intraoperative}. This high failure rate is associated with intraoperative challenges such as unclear tumor boundaries and high tissue mobility. Tumor localization is achieved through palpation, radiological methods such as ultrasound, or wire localization and sketches \cite{athanasiou2022comparative}. However, each of these methods suffer from various shortcomings as  breast tumors can be non-palpable, accessing a radiologist during a surgery is costly and impractical, and interpreting a localization wire sketch is difficult.

To address these limitations, a navigation system called LumpNav has been previously introduced which uses electromagnetic (EM) navigation and ultrasound to help localize the tumor boundaries in real-time \cite{ungi2015navigated, gauvin2020real}. LumpNav provides surgeons with a visual display that shows the position of their resection tool relative to the tumor during surgery and is currently undergoing clinical trials. Although the LumpNav system has demonstrated its efficacy in reducing excised tissue volume and avoiding positive margins, it remains dependent on the visual interpretation of navigation data which can be challenging and mentally demanding for the surgeon. In other anatomical areas, this challenge has been overcome with virtual fixture (VF) guidance, sometimes referred to as active constraints.

Fixtures or constraints are typically used to enforce a desired behaviour through haptic feedback. In the context of robot-assisted surgery or medical procedures, this desired behaviour can include avoiding delicate or critical anatomy \cite{xia2008integrated}, performing a task more efficiently \cite{marinho2020virtual}, trainee skill development \cite{hong2016haptic}, and preventing undesired forces \cite{ishida2024beyond}. Forbidden region VFs in particular can be used to prevent the operator from breaching an undesired region \cite{abbott2007haptic}. In the past 20 years, there have been several advancements in robot-assisted surgery and medical interventions, and virtual fixtures have remained a core component of emerging research and commercial medical robotic systems. For example, the Mako robot (Stryker, USA) uses a forbidden region VF to limit resection depths and preserve soft tissue during bone sawing for total knee arthroplasty \cite{roche2021mako}. Similarly, the ROSA One robot (Zimmer Biomet, USA) uses forbidden region VFs for trajectory assistance in neurosurgical applications \cite{fomenko2022introduction}. The use of robotics in breast surgery, in general, is also being actively investigated at various clinical centers for mastectomy because of its potential positive impact on minimally-invasive surgery \cite{toesca2017robotic, lai2020consensus}. Despite these commercial successes and the increasing interest in robotic-assistance from breast surgeons, the application of VF guidance in BCS remains under explored. The potential benefit of incorporating VF guidance to prevent tumor breach in BCS warrants careful consideration. However, changing the standard of care so significantly, such that the surgeon will cooperatively manipulate their resection tool with a robot, requires thorough evaluation. This is essential to ensure that such changes do not disrupt established surgical practices and outcomes. We hypothesize that while the integration of VF guidance could be advantageous in BCS, its implementation must be thoughtfully considered to balance added value and practicality.

Therefore, in this work, we demonstrate a forbidden region VF guidance system for BCS that relies on ultrasound and EM tracking. While assessing the potential benefits of this system, we also explore the impact on surgical outcome, the acceptance among users, and the potential indirect effects on surgical workflow. To investigate the utility of this guidance system, we conduct a study where users perform a simulated BCS procedure with and without VF guidance. We demonstrate that haptic feedback reduces positive margins and alleviates mental load; at the same time it may increase resection time and margin width. The insights gained from this pilot study will inform the necessary design adjustments, training protocols, and operational parameters for future implementations of haptic guidance in BCS. Our contributions include: 1) the development of an open-source VF guidance system for soft-tissue tumor resections, 2) a new technique for imposing haptic constraints based on real-time EM tracking and ultrasound, and 3) a comprehensive analysis of user interaction and with cooperative robotic guidance in this surgical scenario.

\section{Methods}\label{sec2}

\subsection{System overview}
\begin{figure}[h]
  \centering
  \includegraphics[width=0.87 \textwidth]{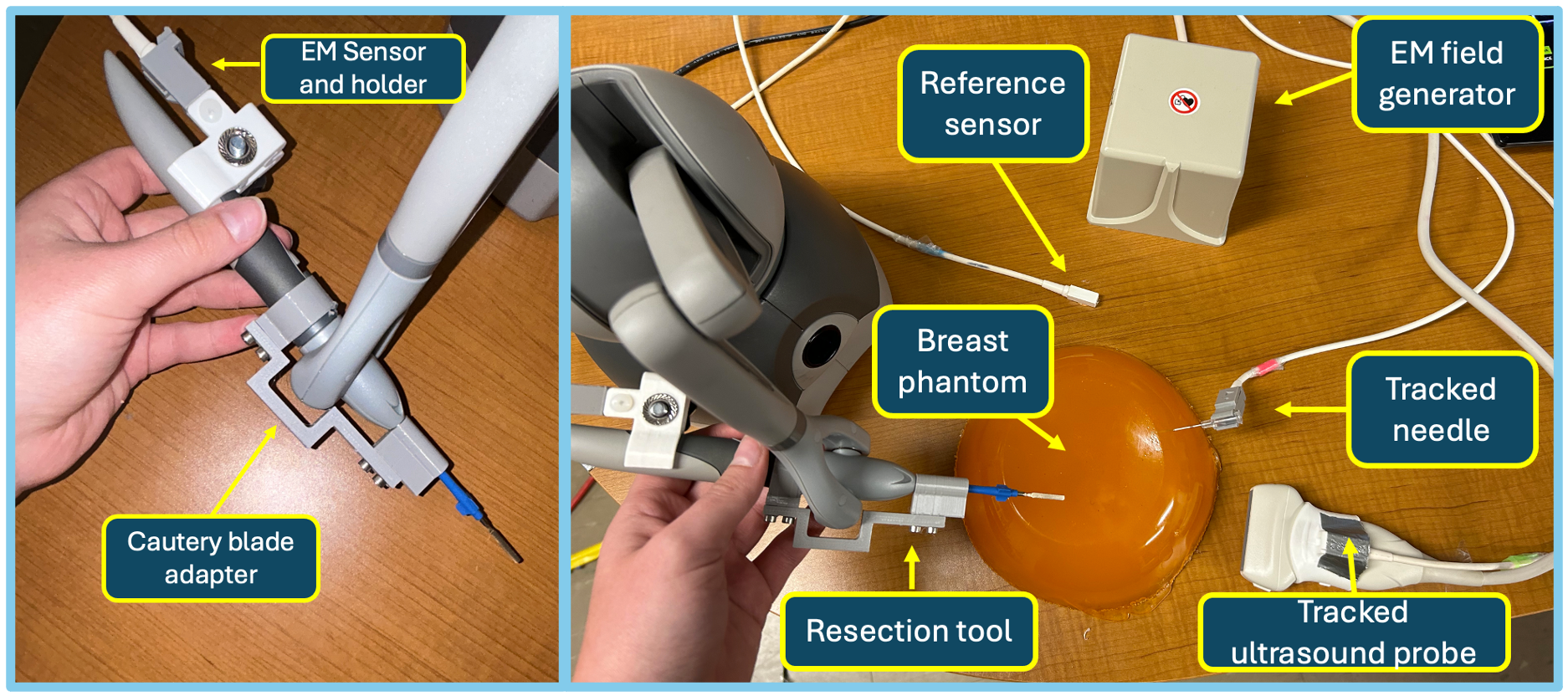}
  \caption{\emph{Left:} Close-up of resection tool which is the stylus of the Touch modified to hold a dull electrocautery blade.  \emph{Right:} The entire system, including the EM field generator, tracked needle, tracked utlrasound probe, reference sensor, and breast simulant.}
  \label{fig:Overview}
\end{figure}

Our proposed guidance system integrates an EM navigation system (Ascension 3DG EM tracker, NDI , Canada), an ultrasound probe (Telemed, Lithuania) and a Touch haptic device - a cooperative robotic interface formerly known as Phantom Omni  (3D Systems). The system is displayed in Figure \ref{fig:Overview}.

Since breast tissue is soft and poses practically no resistant forces on the surgeon's tool, we implement this system with a small haptic device. Although the Touch is sufficient for bench-top testing, in future iterations of this system we envision a robot with a larger workspace for clinical deployment. The device is modified by attaching a blunt electrocautery blade to the tip of the stylus. This is achieved with a 3D-printed adapter that is used to attach the blade to the rotating element of the stylus. An EM sensor is fixed to the stylus of the robot and a needle that is embedded in the tumor before the resection. The embedded needle captures the relative motion of the tumor during the resection. We assume that deformation of the tumor is limited because tumor tissue is generally less elastic than the surrounding breast fat tissue. Both of these sensors are tracked relative to a reference sensor that is fixed to the table. It should be noted that the interaction of the EM tracking field and the Touch robot was evaluated in \cite{connolly2021open} and it was determined that the robot does not distort the magnetic field or negatively influence EM tracking.

\subsection{Integration}

To deploy our virtual fixture, we rely primarily on two computer programs: Asynchronous Multi-Body Framework (AMBF) \cite{munawar2019real} and 3D Slicer \cite{pieper20043d}. AMBF is a real-time dynamic simulation engine that runs both physics computations and graphics in efficient separate threads (\url{https://github.com/WPI-AIM/ambf}). 3D Slicer is an open-source medical imaging platform that is used for surgical navigation (\url{slicer.org}). We rely on AMBF for collision detection and haptic feedback, and 3D Slicer for visual navigation, fixture definition from ultrasound, and tool calibration. Communication between the hardware and software used in this system is facilitated with PLUS toolkit \cite{lasso2014plus},  OpenIGTLink \cite{tokuda2009openigtlink}, Robot operating system (ROS) \cite{koubaa2017robot}, and SlicerROS2 \cite{connolly2024slicerros2}.

\begin{figure}[h]
  \centering
  \includegraphics[width=1.0\textwidth]{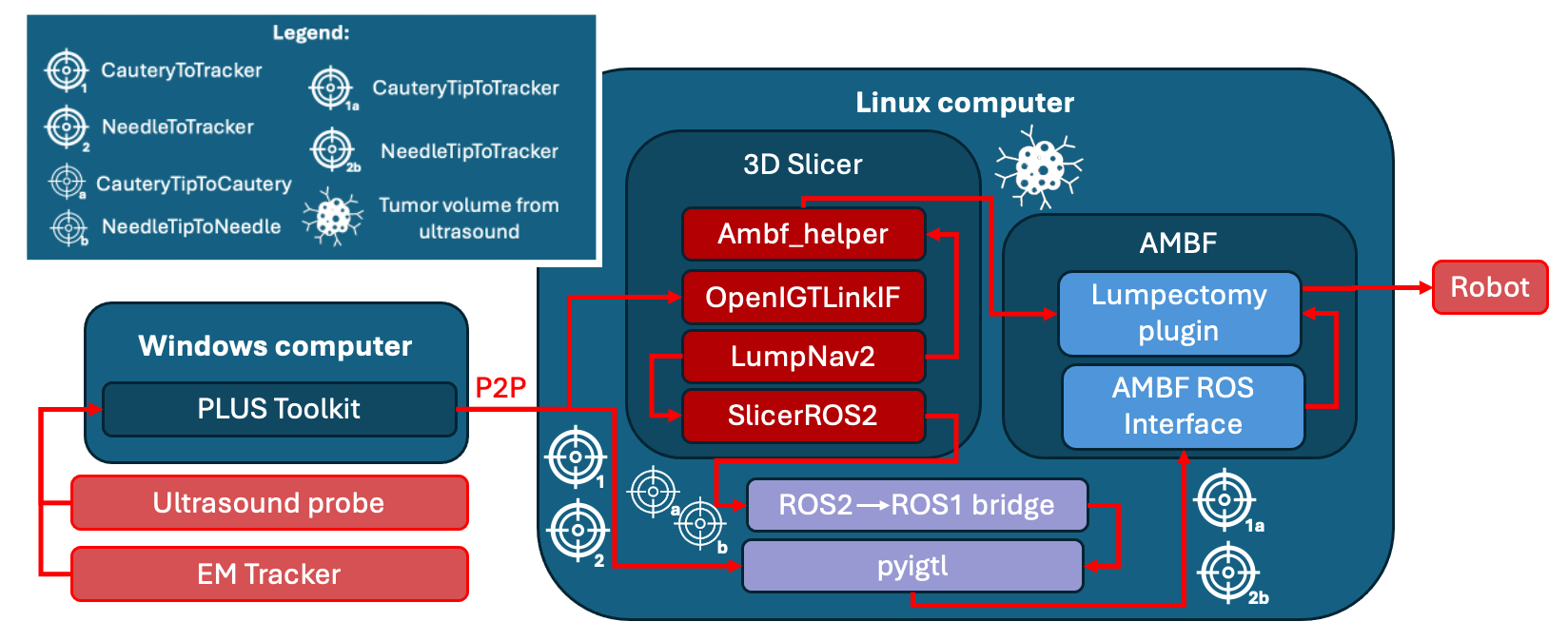}
  \caption{Software and hardware integration. In the top left legend the symbols associated with transformations that are sent between each program are also defined. Cautery in this diagram refers to the modified robotic resection tool.}
  \label{fig:integration_scheme}
\end{figure}

The drivers for the ultrasound probe and EM tracker are only available for a Windows operating system, whereas the rest of the application runs on Linux. We overcome this incompatibility using a simple peer-to-peer network. We also separate the navigation system from the collision detection and haptic feedback for computational efficiency. With this integration scheme, we can run the collision detection and haptics at a rate of 1000~Hz, independent of the visual navigation display. Figure \ref{fig:integration_scheme} illustrates this integration scheme.

\subsection{Fixture definition}

\begin{figure}[ht]
  \centering
  \includegraphics[width=1.0\textwidth]{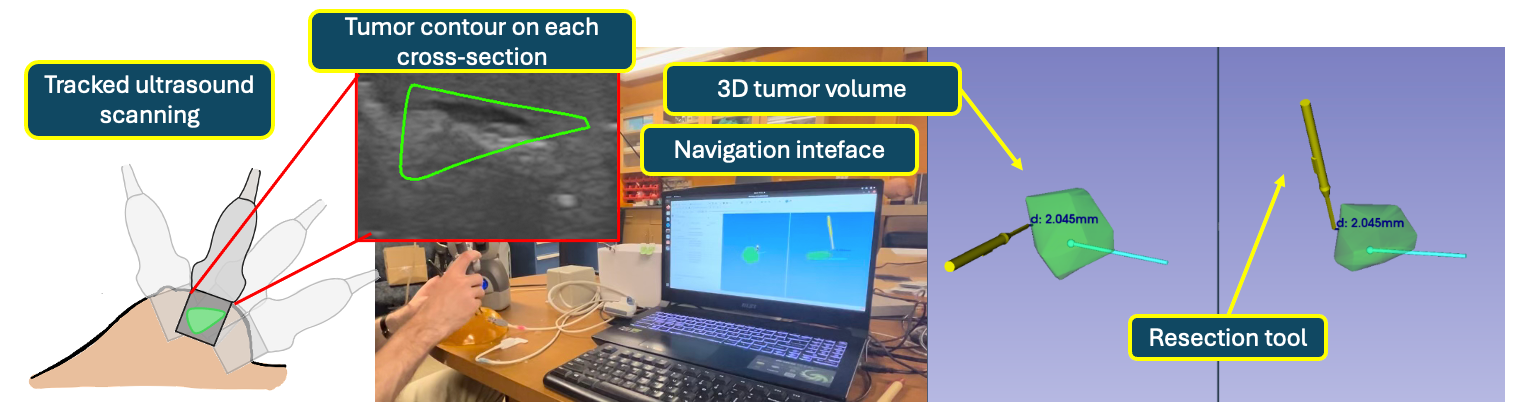}
  \caption{\emph{From left to right:} Visual of the ultrasound motion during contouring. An example of the tumor contour on an ultrasound slice and the experimental set up with the robot and navigation interface on a laptop. Navigation views provided to the surgeon, frontal and sagittal view respectively.}
  \label{fig:tumor definition}
\end{figure}

We use the LumpNav module for visual navigation \cite{gauvin2020real}. Within LumpNav, we do pivot calibration of the tracked needle and resection tool. The needle is first inserted into the tumor under ultrasound guidance. The tracked ultrasound probe is then used to annotate the tumor boundaries. Each of these contours is meant to capture the visible tumor and a small margin of healthy tissue around this boundary. This contouring follows a process, where several cross sections of the tumor are used to annotate the tumor boundaries in 2D. On average, the operator would segment the tumor on approximately 15 cross-sections. The output of these 2D annotations is encapsulated in a convex hull which is assumed to capture the extent of the tumor in 3D. During the experiment, the user is given a frontal (top-down) and sagittal (side) view of their tool relative to this tumor volume. This process is illustrated in Figure \ref{fig:tumor definition}.

After the contour is generated and exported to AMBF using the pipeline highlighted in Figure \ref{fig:integration_scheme}, the VF guidance is initialized. For haptic feedback, we use the CHAI3D finger proxy collision algorithm previously shown in \cite{munawar2022virtual}. The tip of the resection tool is approximated as a spherical proxy that cannot penetrate the tumor in the AMBF environment. When this contact point collides with the tumor volume, this position at the tumor surface is captured as the ``proxy" and the actual pose of the operator's tool becomes the ``goal". A vector between the goal position and the proxy position is then used to compute a proportional resistant force that is sent to the Touch. This collision algorithm is further outlined in \cite{ruspini2001haptic}. For this experiment, the proxy has a collision margin of 4 mm. This size was selected to enforce a conservative margin that meets the accepted clinical standard of 2 mm \cite{bundred2022margin} while accounting for the $\sim$1-2 mm accuracy of the EM tracker. It was necessary to use the maximum force range of the device (up to 3.3 Newtons) to ensure the haptic feedback was still detectable when cutting through the breast simulants.

\subsection{Breast cancer simulants}
There is no accurate animal model for breast tissue that can be used to simulate a surgical procedure. Similarly, breast tissue undergoes tissue necrosis shortly after it has been resected from the patient which makes it infeasible to perform cadaver experiments. Considering these challenges, we have developed gel wax breast simulants that can be used to simulate a surgical excision of breast tumor tissue. The healthy tissue is made by melting medium and low density gel wax (Virginia Candle Supply, Tennessee, USA), then mixing in wax dye and cellulose for ultrasound scattering. The tumor tissue is made with gel wax, and a small amount of epoxy resin luminous powder (Limino, Guangzhou, China). Breast tumors are generally 2-5 cm in diameter so we used a circular mold with a diameter of 3.5 cm to cut the tumors and ensure that they are all uniform width and appropriately sized. Simulants were constructed by pouring gel wax into a metal bowl up to the halfway point, letting this material cool until the surface had hardened, adding in the tumor, and then pouring more gel wax on top to seal the tumor inside. The resulting simulants had tumors that were indistinguishable in regular lighting but visible under ultraviolet (UV) light and in ultrasound. Controlling the position of the tumor and the height of the simulants was challenging due to the nature of the materials so each simulant had some uniqueness with respect to the depth and position of the tumor and the final shape once the gels had mixed and cooled. This variability is similar to a clinical scenario where the visibility of the tumor in ultrasound, shape, and extent would vary significantly from patient to patient.

\subsection{Experimental protocol} 

To evaluate the effectiveness of our VF guidance system, six users who were familiar with surgical navigation were asked to remove a tumor from breast simulants using our modified resection tool, both with and without haptic feedback. Two of the users are medical trainees, one user is an attending radiologist, and three of the users are engineering / computer science trainees, specializing in computer assisted surgery. In both scenarios the users were given the same navigation view that is available with LumpNav (illustrated in Figure \ref{fig:tumor definition}). Before the resection, each user was instructed to watch the same video that included a clinical example of breast-conserving surgery, a video explaining the LumpNav navigation system, and the details of the experiment. They were also each allowed a few minutes to use the resection tool to cut the gel wax in a practice simulant so they could become comfortable with the sensation of cutting. Users were asked to perform six resections, in three rounds with a break in between each round. Each user was instructed to use the robot guided resection tool to use to first define the tumor boundaries. After the borders were defined, they were also given a scalpel and tweezers to help remove the specimen. They were instructed to cut right through each simulant. The first two rounds were used to get familiar with the task and the third was used for evaluation. By the time the user got to the final evaluation round, they had completed four resections in total, two with VF guidance and two without. The order of the experimental scenarios, i.e. starting with or without haptic feedback during each trial,  was alternated and assigned randomly to each user to avoid the influence of the learning effect on our results. In the end, half of the users were given VF guidance first and the other half were given VF guidance last.

\begin{figure}[ht]
  \centering
  \includegraphics[width=0.94\textwidth]{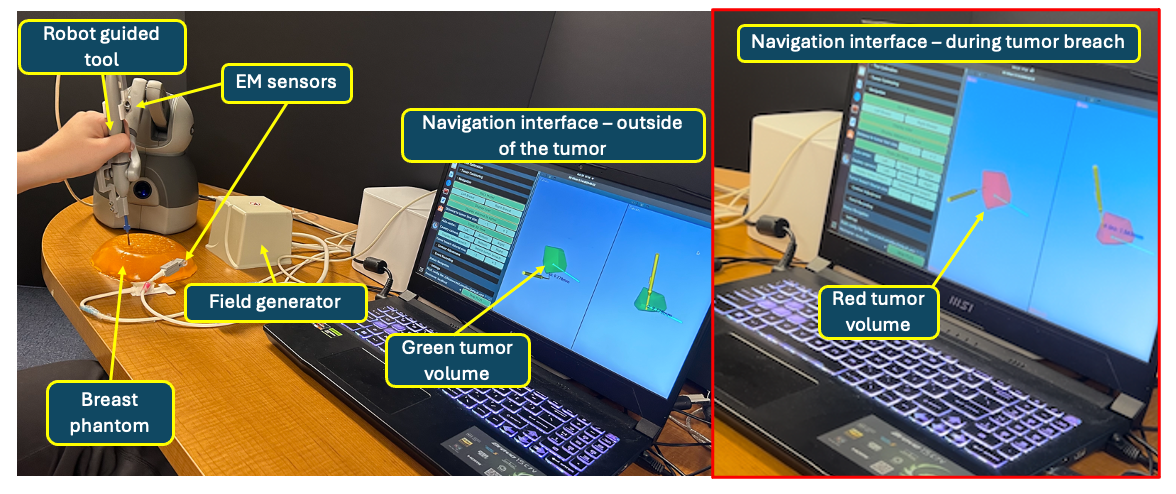}
  \caption{\emph{Left:} User performing experiment with VF guidance system. Navigation interface is shown when the tip of the resection tool is outside of the tumor. \emph{Right:} Navigation interface is shown when the the tip of the resection tool is in contact with the tumor.}
  \label{fig:my_label}
\end{figure}

To objectively evaluate the performance, we calculated the total resection time and the percentage of the total volume that was removed across all 24 resections. The total resection time is considered the full length of the procedure from first incision to complete excision and was timed using the sequences module in 3D Slicer. The percentage of total volume removed was calculated with a scale and was determined by measuring the weight of the removed specimen divided by the weight of the full simulant to avoid any bias introduced by different sample sizes. The resection margins were also evaluated after each resection using a UV flashlight in a dark room.

After the experiment, the users were also asked to fill out a NASA Task Load Index (TLX) survey to better understand their experience with the guidance system \cite{hart2006nasa}. The performance question from the NASA TLX survey was posed as to how successful the user felt they were in completing the task as they had no indication as to how well they actually did when filling out the survey. The NASA TLX scale goes from 0 to 21 where the lower end indicates less effort and the higher end indicates more effort. This serves as our subjective evaluation.

\section{Results \& Discussion}\label{sec12}

The final resections from the evaluation round are highlighted in Figure \ref{fig:Vf-revisedResults} below. The top row is the resections without any VF guidance, ie. visual navigation only, and the bottom row is the resections with VF guidance. For each user, we show the resected specimen that was removed from the simulant as well as the complimentary resection cavity that was left behind. The presence of blue dye in the resection cavity indicates a positive margin. 

\begin{figure}[h]
  \centering
\includegraphics[width=0.99\textwidth]{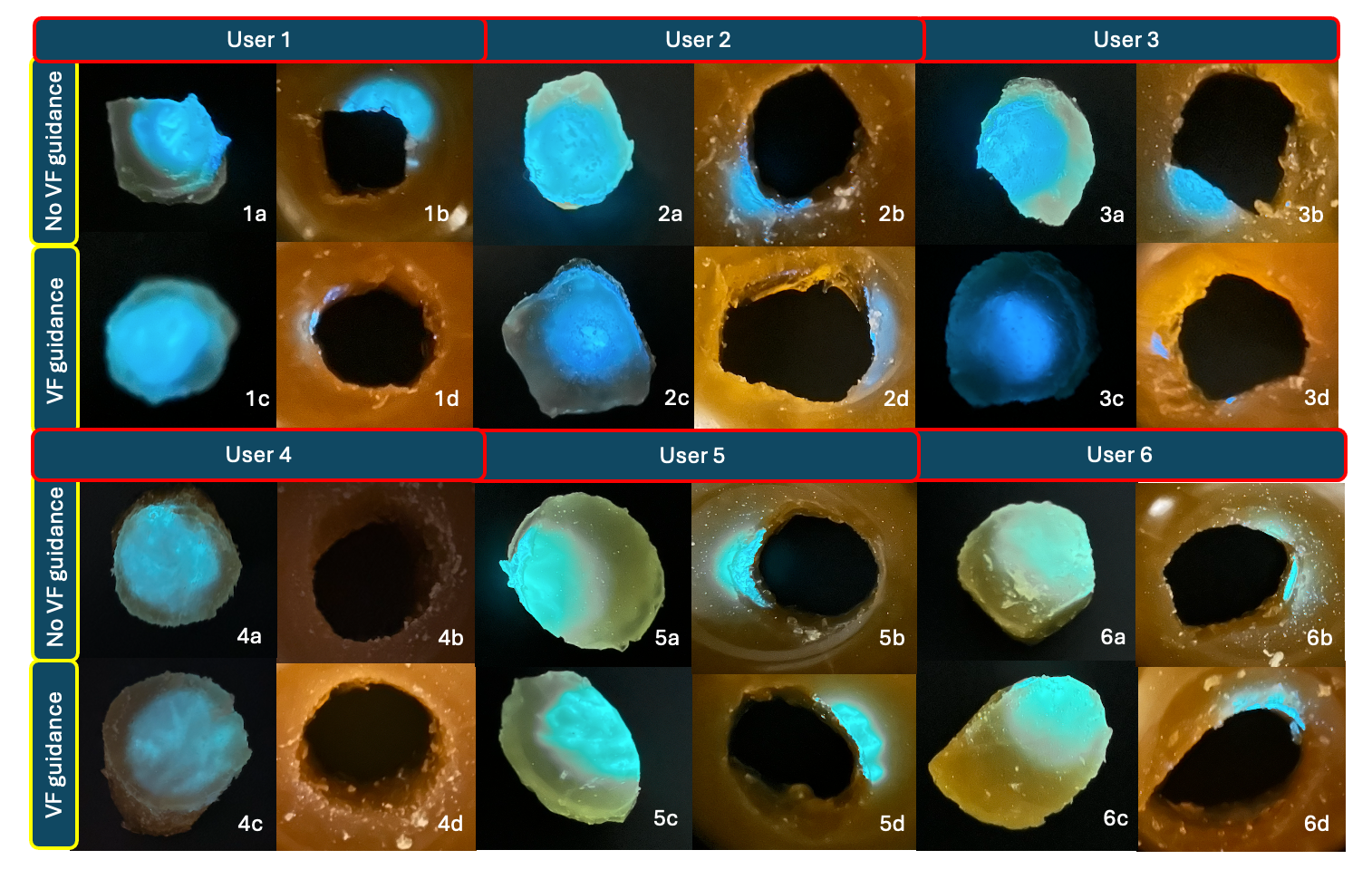}
  \caption{\emph{Left to right:}Final resection, third round, from each independent user. The top row for each user is resections without VF guidance and bottom row is resection with VF guidance. For each user, the images in the left block show the resected specimen and the images in the right block show the complimentary resection cavity in the simulant. Blue in the resection cavity on the right indicates the presence of a positive margin. \emph{Top to bottom:} Results from users 1-3, results from users 4-6.}
  \label{fig:Vf-revisedResults}
\end{figure}

Generally users left behind less residual tumor tissue or no positive margins when the VF guidance was active versus when they were relying on visual guidance alone. This is particularly evident for users \#1 and \#3, where the resection cavities contain much less residual tumor tissue (Figure \ref{fig:Vf-revisedResults} - 1d, 3d) than their visual navigation counterparts (Figure \ref{fig:Vf-revisedResults}- 1b, 3b). In fact, the presence of positive margins in these cases appear to be caused by satellite tumor tissue that may have been indistinguishable in ultrasound. As opposed to 1b and 3b, where it appears that the user transected the tumor and left behind substantial residual tumor. User \#2 leaves behind a similar resection margin when VF guidance is active or inactive, although the margin that is left behind with VF guidance appears to be the result of tight delineation of an accurate boundary rather than inaccurate delineation. User \#4 appears to achieve the same negative margins both with and without robotic guidance. A similar outcome is observed for users 5 and 6 where both users leave behind very similar positive margins both with and without haptic feedback. Although these results are visually distinct in Figure \ref{fig:Vf-revisedResults}, in future studies we will incorporate computational measurement techniques to quantify these margins.

The average scores from the NASA TLX survey from each user are highlighted in Figure \ref{fig:barchart-NASA} below. 

\begin{figure}[ht]
  \centering
\includegraphics[width=1.0\textwidth]{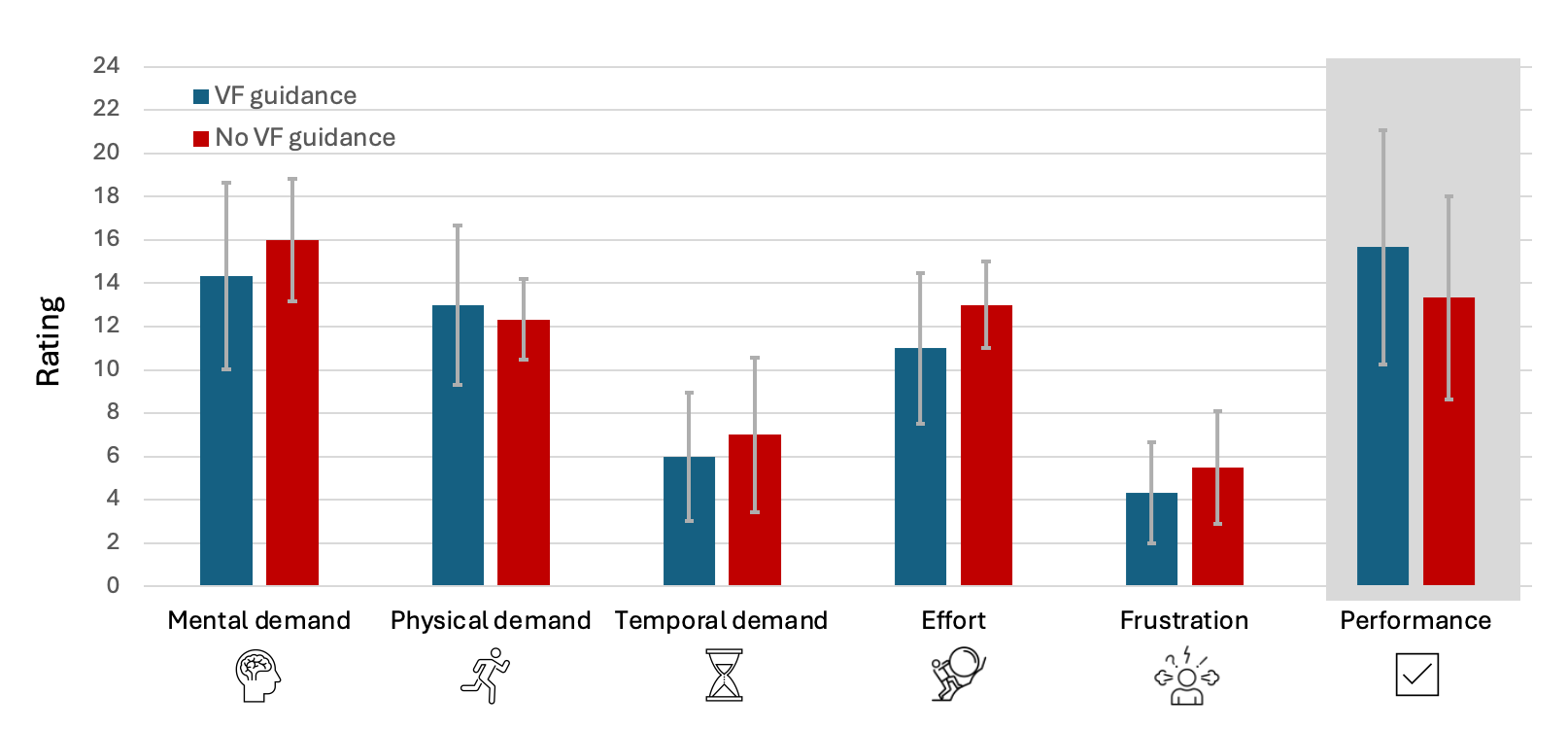}
  \caption{Average NASA scores across all six users. From left to right: Mental demand, physical demand, temporal demand, effort, frustration and percieved performance. Error bars indicate the standard deviation across individual scores.}
  \label{fig:barchart-NASA}
\end{figure}

These results indicate that, on average, users felt less mental and temporal demand as well as less effort and frustration when using the haptic feedback. The physical demand was considered higher for the haptic feedback which is expected because the user is not only managing resistant forces from the breast simulant, but also the haptic feedback from the device. The majority of users also felt more confident in their performance when they had the VF guidance versus when they did not (last 2 bars in Figure \ref{fig:barchart-NASA}) which is also consistent with our results shown in Figure \ref{fig:Vf-revisedResults}.

Finally, Table \ref{Objective results} below summarizes the percentage of total volume removed and the resection time in minutes for each experimental condition. These metrics are calculated for the training resections and for the final resections that were obtained in the evaluation round only. The column ``VF guidance" indicates that the robot was on and providing haptic feedback and the column ``No VF guidance" indicates that the robot was off.

\begin{table}[h!]
\centering
\caption{Average resection time and percentage of total volume removed during experiments.}
\begin{tabularx}{\textwidth}{X X X X X} 
\toprule
\textbf{Average} & \multicolumn{2}{c}{\textbf{Training rounds}} & \multicolumn{2}{c}{\textbf{Evaluation round}} \\
 & VF guidance & No VF guidance & VF guidance & No VF guidance \\
\midrule
Resection time & $13.8 \pm 6.7$ mins & $9.5 \pm 3.1$ mins & $13.9 \pm 5.4$ mins & $9.8 \pm 4.0$ mins \\
\midrule
\% of total volume removed & $16.3 \pm 3.9$ \% & $14.4 \pm 5.4$ \% & $17.5 \pm 3.5$ \% & $15.2 \pm 1.4$ \% \\
\bottomrule
\end{tabularx}
\label{Objective results}
\end{table}

These results indicate that the resection time is generally longer when using VF guidance. This can be attributed to a pattern that was observed during the user study where participants would spend more time palpating the tumor boundaries before removing the specimen when they had access to haptic feedback. In contrast, those using only visual guidance tended to rely on their initial resection outlines. Therefore, while VF guidance typically resulted in less residual tumor tissue, it also introduced a trade-off between time spent and accuracy of tumor delineation. This is a critical design consideration, as the cost of leaving behind a positive margin and requiring a revision surgery far outweigh the extra time that may be spent palpating in the first procedure. Importantly, as users gained experience, their efficiency improved, suggesting that the observed time increase might diminish with adequate training.

We also observed that the percentage of total volume removed is higher for the cases with VF guidance. This was generally beneficial as it often resulted in a more complete tumor resection. In instances where positive margins did occur when VF guidance was active, they were also largely due to tight tumor delineation rather than inadequate tissue removal. To this end, future investigations will focus on optimizing the placement of forbidden region boundaries and enhancing haptic transitions between healthy and tumor tissues to standardize resection margin widths more effectively. We will also consider incorporating soft tissue modeling to reflect potential changes and deformations to the tumor tissue, and consequently forbidden region boundaries. Considering this system is implemented with open-source software like 3D Slicer and ROS, we will follow the integration of this technology into these platforms.

One of the limitations of this study is that the tumor boundaries and virtual fixture constraints were delineated manually in tracked ultrasound which is heavily reliant on the visibility of the tumor in ultrasound and on the operator. Although this is consistent with the current workflow that is employed in LumpNav clinical trials, and the same operator contoured the tumors for each experiment, we noted that some of the simulated tumor would expand after the second layer of the gel was poured which made these boundaries difficult to identify manually in some cases. This is an existing clinical problem that is currently being investigated with AI tumor segmentation models \cite{yeung2024quantitative}. In future work, these models will be incorporated into this pipeline to ensure that the tumor contours capture the entire tumor and boundaries that may not be visible to a human operator. 

We did not compare the resections with VF guidance to a freehand baseline. This decision was made to prevent the workspace limitations of the Touch device from influencing the user's perception of VF guidance. We imagine a device with a larger and less confined workspace will be necessary for clinical adoption therefore, in future work, we will also include a freehand baseline in our evaluations.

\section{Conclusion}\label{sec13}
 
This pilot study underscores the potential of VF guidance in BCS to improve surgical outcomes by reducing positive margins and potentially easing the surgeon's cognitive load. However, it also highlights the necessity for further refinement of VF system design and the importance of comprehensive training to fully realize these benefits while mitigating unintended impacts on surgical workflow and outcomes. Future work will explore how different feedback mechanisms and training protocols can be optimized to better integrate VF guidance into BCS practices. We will work to ensure that surgeons can both feel and understand how it should feel when they \emph{touch} the tumor boundary.

\backmatter

\section*{Statements and Declarations}

\noindent
\textbf{Funding:} Laura Connolly is supported by an National Science and Engineering Research Council (NSERC) Canada Graduate Scholarship - Doctoral (CGS-D) award and the Vector Institute. Gabor Fichtinger and Parvin Mousavi each hold a Canada Research Chair, Tier 1. Parvin Mousavi also holds a Canada CIFAR AI Chair. Russell H. Taylor is supported by Johns Hopkins University Internal Funds. \\

\noindent
\textbf{Competing Interests:} We have no conflicts of interest to declare. \\

\noindent
\textbf{Ethics and informed consent:} This study was approved by Queen’s University Health Sciences Research Ethics Board. All participants provided informed verbal and written consent. \\

 \noindent
\textbf{Data \& code availability:} The data from this study is not available. The code will be made available on GitHub upon acceptance.

\bibliography{sn-bibliography}
\input{output.bbl}

\end{document}